# AUTOMATED MANAGEMENT OF POTHOLE RELATED DISASTERS USING IMAGE PROCESSING AND GEOTAGGING

Manisha Mandal[1], Madhura Katageri[2], Mansi Gandhi[3], Navin Koregaonkar[4] and Prof. Sharmila Sengupta[5]

[1]Department of Computer Engineering, Vivekanand Education Society's Institute of Technology, Mumbai
*manisha.mandal@ves.ac.in*

[2]Department of Computer Engineering, Vivekanand Education Society's Institute of Technology, Mumbai
*madhura.katageri@ves.ac.in*

[3]Department of Computer Engineering, Vivekanand Education Society's Institute of Technology, Mumbai
*mansi.gandhi@ves.ac.in*

[4]Department of Computer Engineering, Vivekanand Education Society's Institute of Technology, Mumbai
*navin.koregaonkar@ves.ac.in*

[5]Department of Computer Engineering, Vivekanand Education Society's Institute of Technology, Mumbai
*sharmila.sengupta@ves.ac.in*

## ABSTRACT

*Potholes though seem inconsequential, may cause accidents resulting in loss of human life. In this paper, we present an automated system to efficiently manage the potholes in a ward by deploying geotagging and image processing techniques that overcomes the drawbacks associated with the existing survey-oriented systems. Image processing is used for identification of target pothole regions in the 2D images using edge detection and morphological image processing operations. A method is developed to accurately estimate the dimensions of the potholes from their images, analyze their area and depth, estimate the quantity of filling material required and therefore enabling pothole attendance on a priority basis. This will further enable the government official to have a fully automated system for effectively managing pothole related disasters.*

## 1. INTRODUCTION

A pothole is a bowl-shaped depression in pavement surface. It can be caused due to internal factors like pavement erosion by water seeping under it, due to change in climate, like heavy rainfall, and external factors such as poor construction management and heavy traffic. The presence of potholes leads to damage of vehicles, accidents and even death [1] in many cases which also causes many legal complications. Figure 1 shows one such newspaper article. It is therefore necessary to detect, recognize and repair the potholes to ensure reduction in risk to human lives and vehicles due to such non substantial reasons.

The pothole management system adopted by most government agencies is survey oriented, requiring heavy usage of resources like manpower, data collection and data analytics tools and is not foolproof. Also as there is limited automation applied, the probability and chances of getting optimal inferences is highly reduced.

The objective of this paper is to introduce a system that uses image processing techniques on pothole images. The system aims to provide statistical information as well as a final geotagged image of an area with pothole location and information in a fully automated manner, with priority based complaint attendance having the added benefit of efficient utilization of materials in an extremely economical way. This would accordingly provide the concerned government body with the various information that they might require to manage the potholes in that area.

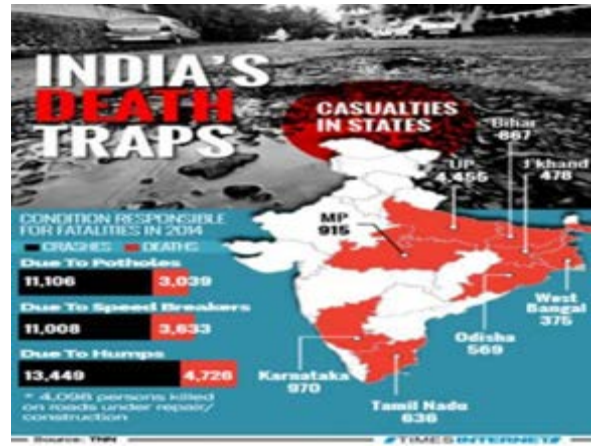

Figure 1. Newspaper article highlighting death rate of casualties due to potholes

## 2. LITERATURE SURVEY

### 2.1. Related Work

Existing pothole detection methods [2] can be divided into vibration-based methods, 3D reconstruction-based method and vision-based methods. Vision-based methods are of two types; 2D image-based approaches by Koch and Brilakis [3] and Buza et al. [4] and video-based approaches. Vision-based techniques are cost effective as compared to 3D laser scanner methods. However due to noise, distorted signal is generated in the case of vision based methods and thus we need to develop and improve existing detection methods.

Koch and Brilakis presented a supervised method for detecting potholes in an automated way with an overall efficiency of 86% [3]. Under the proposed method, the image is segmented into defect and non-defect regions. The potential pothole shape is then approximated according to the geometric characteristics of a defect region. Next, the texture of a potential region is extracted and compared with the texture of the surrounding region. This implies that the system is trained from a number of texture samples. In order to emphasize structural texture characteristics, the grayscale image is passed through four spot filters. This method assumes all images are shot from same distance and angle and thus no scaling of absolute filter size has been done. However, the images taken from varying heights, angles and proximity need to be considered.

Buza et al. [4] proposed a new unsupervised vision-based method consisting of three steps such as image segmentation, shape extraction and identification and extraction. Histogram based thresholding is used for image segmentation.Instead of using traditional clustering algorithms like k-means algorithm, normalized spectral clustering algorithm have been used for shape

identification.Seeds are then selected which helps to extract vertical and then horizontal area.81% accuracy was obtained for estimation of a pothole surface area.This method works on pictures taken from good perspectives with explicit focus on the road anomaly,which is not usually possible in real time scenarios.

Rajab et al. use ImageJ software for image processing to determine areas of a potholes and show that results from image measurement methods, which are safe, fast and effortless are close to those obtained by applying traditional methods [5]. Their method uses curve fitting to specified points at the pothole border to measure the area of a pothole. However, the images are acquired with necessary road marks and units of known length like metal rulers. Such images would be difficult to obtain for all the potholes.

Nienaber et al. [6] used a GoPro camera attached to the front windscreen of a car. The extracted road images were converted to grayscale images and Gaussian filter was applied to remove noise. Canny edge detection was performed followed by dilation to remove unwanted edges close to outer boundaries. Sample study performed by them indicated a precision of 81.8% and recall of 74.4%. Pawade et al. [7] proposed a FGPA(Field Programmable Gate Arrangement) system to detect potholes because of its easy reconfiguration and rewritable logic ability that enables the detection system to change its logic every time a new image of the road is captured by the camera. The processing part of the system included detection of potholes using basic edge detection algorithms like Prewitt, Sobel and Canny. The cost overhead generated by the installation of these camera equipments in both the above methods acts as a disadvantage for them.

## 2.2. Proposed System

The proposed system acquires potholes images through camera. Using latitude and longitude information from the images, potholes are geotagged on the selected ward [8] map. As the acquired images would be taken under different environmental conditions, various noise filtering techniques would be employed followed by appropriate edge detection algorithms like Canny and Zerocross to identify the boundary of the pothole in the image. After the target area is determined in the image, its dimensional information like area, size and volume would be determined. For this, as the images are taken from different elevations and angles, the top view of the pothole image would be identified from the available image. Orientation could be changed using transformation algorithms like affine transforms. This would be followed by applying appropriate scaling factors to facilitate the area calculating process. Also metrics need to be developed for the same.

## 3. METHODOLOGY

Even though the other traditional systems for pothole detection involving vehicles equipped with sensors and camera provide accurate results, but the cost overhead associated with them makes them infeasible. This paper aims at introducing a system that is highly inexpensive and provides accurate results with maximum precision. This system as depicted in Figure 2, is broadly divided into three parts- Geotagging, Image Processing and verification of data acquired by a government body.

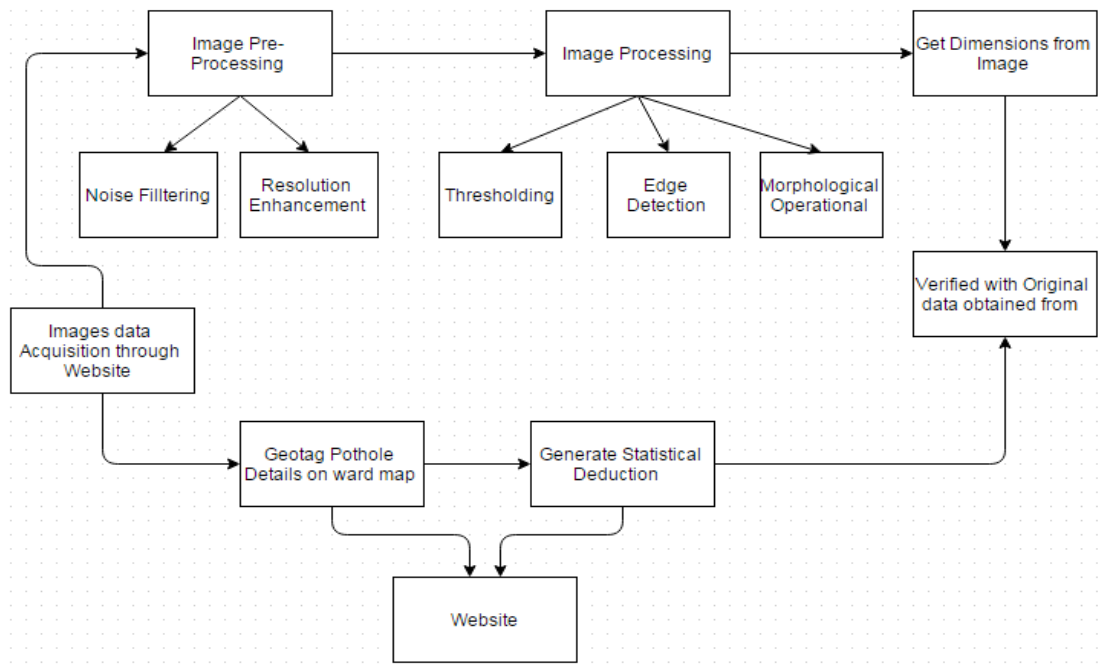

Figure 2. System Design

## 3.1. Geotagging

For Geotagging, the pothole images and data acquired from Brihanmumbai Municipal Corporation (BMC) through their website voiceofcitizen.com [9] had to be consolidated into a centralized database. The images were then geotagged using their latitude and longitude information on Google Earth [10], an open source software. The selected ward [8] was drawn on the map. Further division into sectors was also made. Figure 3 depicts the ward map with sectors marked on it. Google Earth was chosen as it helps in easily geotagging pothole images using placemarks as well as in adding description of the potholes below their images. The ward map with potholes tagged on it using placemarks is depicted in Figure 4. On clicking at the tagged places, the pothole images could be seen along with their exact location as well as description like area and depth as shown in Figure 5. The tagged information is stored in a kmz or kml file that could be easily transferred via emails.

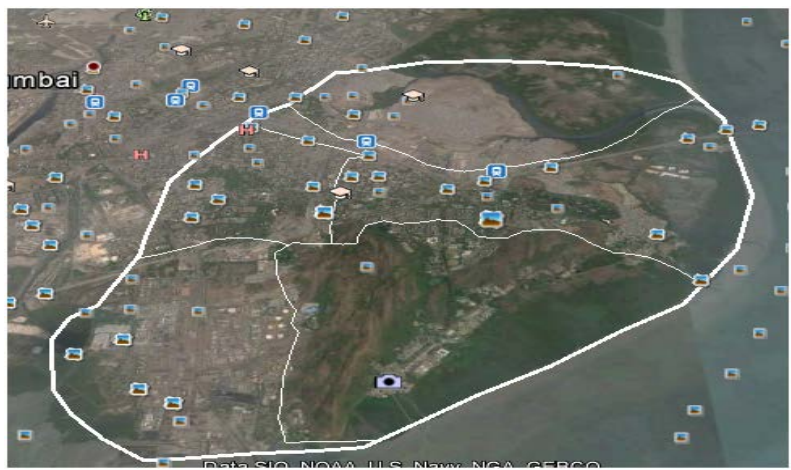

Figure 3. Ward map with marked sectors

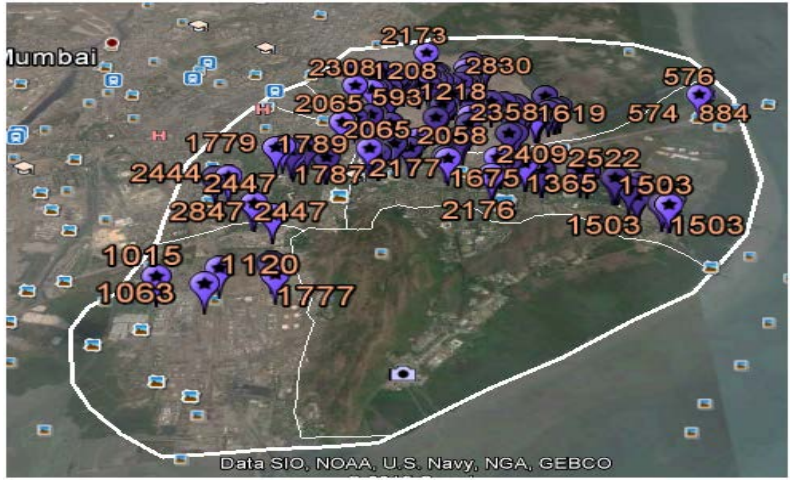

Figure 4. Ward map with potholes geotagged using placemarks.

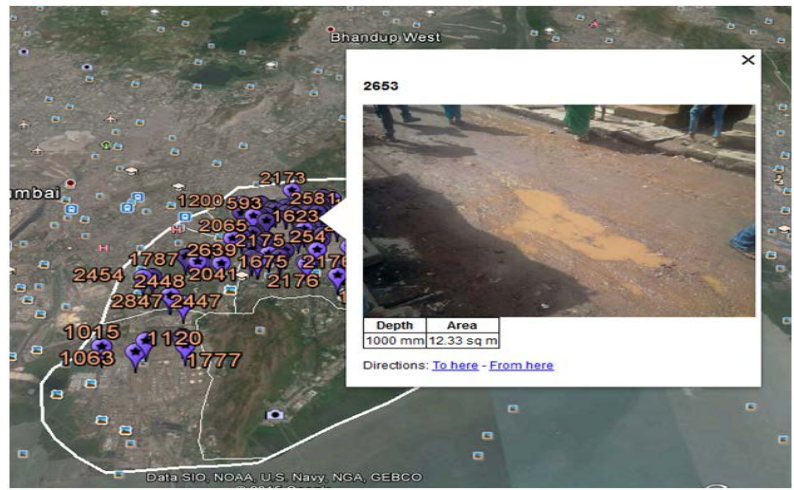

Figure 5. Pothole image and its information.

## 3.2. Image Processing

The input image i(x,y) shown in Figure 6 was first converted to binary image b(x,y) as seen in Figure 7 using the im2bw() function of MATLAB by applying the following formula:

b(x,y) = 1,   if   Y( i(x,y)) >   level
      = 0,    if   Y(i(x,y))  <= level

where,
Y(i(x,y)) represents luminance of the pixel value of the input image and level is a value between [0,1] that is passed as a parameter to the im2bw() function.

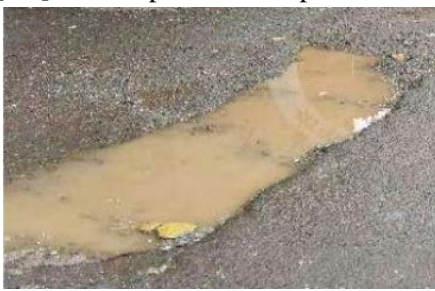 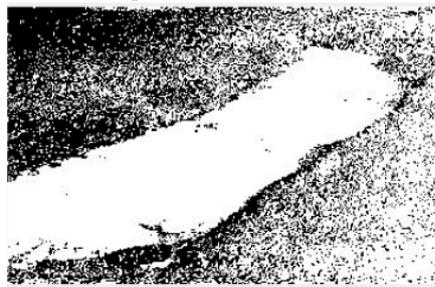

Figure 6. Original Image         Figure 7. Binary Image

Depending on the value of 'level' we get different results of binary images. The original image shown in Figure 8 is converted into binary with the value of 'level' as 0.3, 0.5 and 0.8 as shown in Figure 9, 10 and 11 respectively.

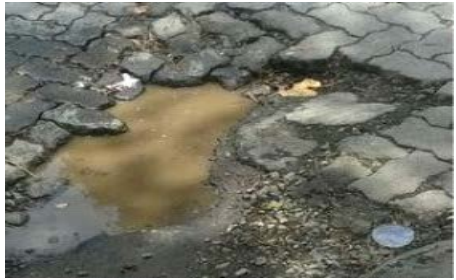
Figure 8. Original Image

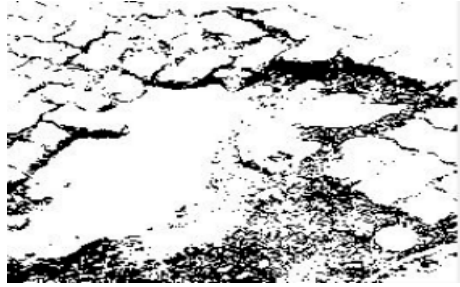
Figure 9. Image with level=0.3

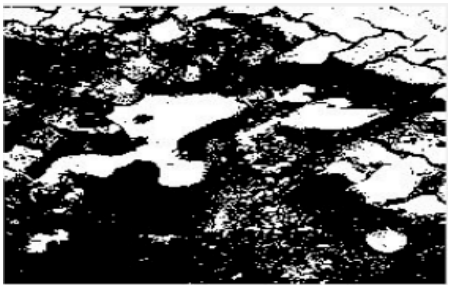
Figure 10. Image with level=0.5

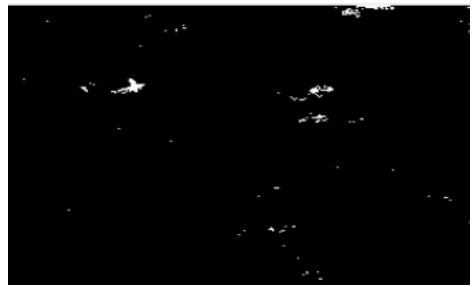
Figure 11. Image with level=0.8

Edge detection algorithms like Canny, Sobel, Prewitts, Roberts and Zerocross were then applied to these binary images. Canny edge detection algorithm provided optimal detection of objects in an image; even the low intensity edges of Figure 8 were detected easily using this algorithm as seen in Figure 12. However in certain images like Figure 6, Canny could not efficiently determine the pothole edges because of same intensity pixel values of the pothole and the plain road area. The result of applying Canny on Figure 6 is shown in Figure 13. Thus for such images Zerocross edge detection algorithm was used, which is based upon the intensity values at the edge pixels. Depending on the change in these intensity values appropriate edges are marked off. Figure 14 illustrates usage of Zerocross edge detection algorithm on the image.

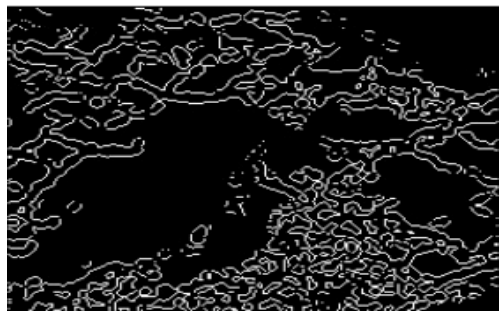
Figure 12. Output of Canny Algorithm

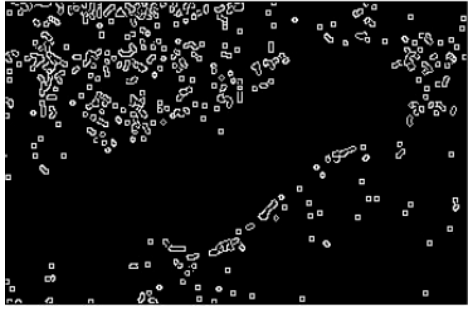 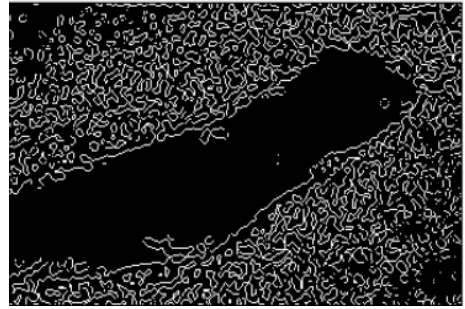

Figure 13. Output of Canny for Figure 8

Figure 14. Output of Zerocross for Figure 8

Once the edges were detected, a structure element of type disk of a particular radius was created. Image closing operation was performed on the intermediate edge detected image and the structured element. The image then obtained was simply inverted so as to specify the main target in white. But in case of certain images the intensity values of the background pixels were less than that of the main target area. In such cases there was no need of inverting the image to highlight the target area. Figure 15 shows one such case. The result of closing operation for Figure 15 is shown in Figure 16. As mentioned above, the target area is already highlighted without having to invert the image. The holes that still existed inside the target area depicted in black, were filled using the imfill() function in MATLAB.

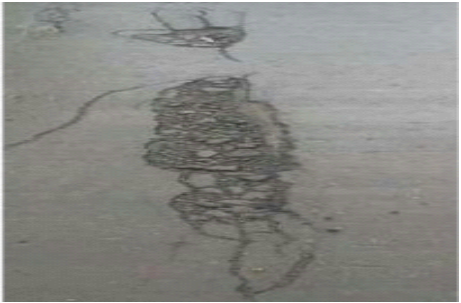 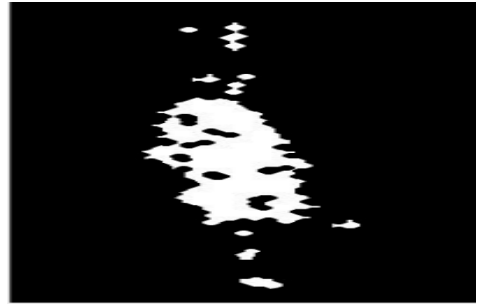

Figure 15. Original Image

Figure 16. Pothole area in white

Finally a RGB image is generated with the pothole area highlighted in blue using the label2rgb() function of MATLAB that uses our label matrix to color all objects in the image.

### 3.3. Verification of data

The third and final part of the system involves verification of data provided by Brihanmumbai Municipal Corporation (BMC) to determine the precision and accuracy of the proposed system. Also priority based pothole attendance report would be generated from this data.

## 4. IMPLEMENTATION AND RESULTS

### 4.1. Implementation

Images of potholes and their data have been acquired from BMC through their website voiceofcitizen.com [9]. As the images need to be geotagged, latitude and longitude information is required along with the images, which could be obtained through a GPS enabled camera. For geotagging pothole images and their information, Google Earth has been used. It provides flexibility as it is free, available for personal computers and mobile viewers and is also available as browser plugin. Placemarks have been used to geotag the potholes along with their description. The tagged file could be easily shared and opened on any machine with Google Earth installed. Image processing techniques on the pothole images presented in this paper have been implemented in MATLAB utilizing the embedded Image Processing Toolbox. Various graphs and pie charts were created using analytical tools like Google Sheets to analyze the data like the number of days to repair the potholes, their volume, etc. These statistical deductions help in determining the priority to be assigned to the repairing of the potholes. This helps in graphically visualizing the data that was present in the theoretical form.

### 4.2. Results

After geotagging the pothole images and generating a centralized database, images were processed to detect the edges of potholes. Figure 17 shows the original image obtained through a GPS enabled camera. This input image was first converted to binary image using the im2bw() function of MATLAB resulting in a threshold image as shown in Figure 18. Zerocross edge detection algorithm was used resulting in Figure 19. Then a disc type structure element was created and closing operation was performed resulting in Figure 20. Inversion was carried out with the pothole area marked in white as shown in Figure 21. Black color holes in the pothole areas were removed using imfill() function in MATLAB. This resulted in Figure 22. Figure 23 shows result of label2rgb() function with the pothole area highlighted in blue.

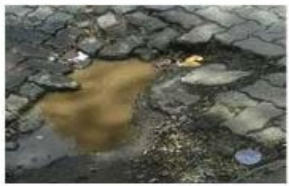
Figure 17. Original Image

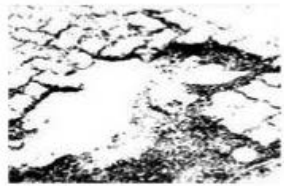
Figure 18. Binary Image

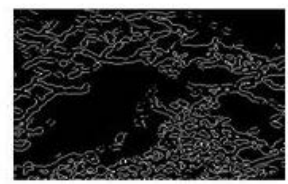
Figure 19. Output of Zerocross edge detection

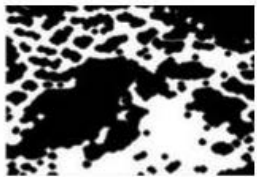
Figure 20. Result of Image Closing

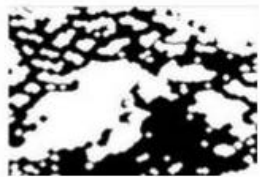
Figure 21. Inversion of the image.

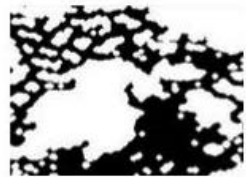
Figure 22. Removal of holes using imfill().

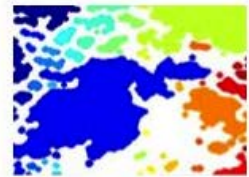
Figure 23. Pothole area in blue, using label2rgb().

Statistical deductions were done using analytical tools like Google Sheets. Figure 24 shows that the number of potholes having area in the range 8 to 13 sq. m. (medium range) was found to be maximum. Also, as compared to small(0 to 50 mm) and large(100mm and greater) depth potholes, the number of medium depth(51 to 99 mm) ones were very high as depicted in Figure 25. Figure 26 shows that there were no potholes with large(14 sq. m and greater) area and small depth in the selected ward. It was seen that almost all potholes were attended within 23 days as illustrated in Figure 27. Pie chart in Figure 28 depicts that the number of potholes having medium area and medium depth were highest compared to other categories.

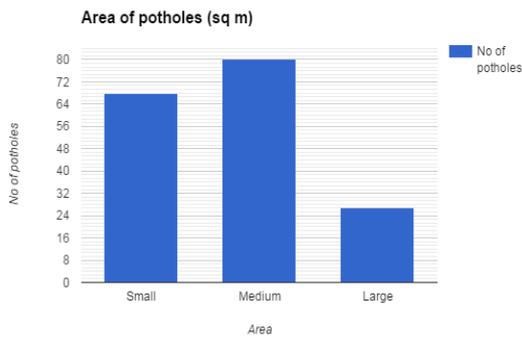 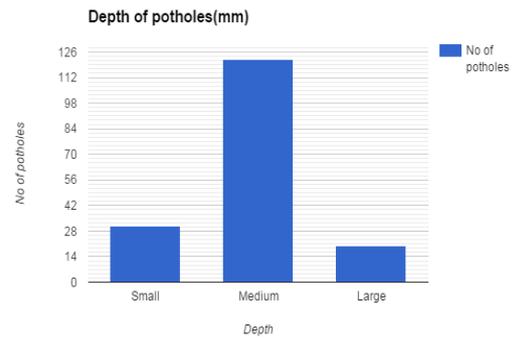

Figure 24. Number of potholes vs area     Figure 25. Number of potholes vs depth

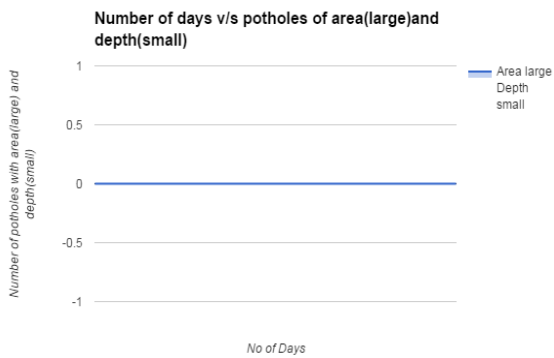 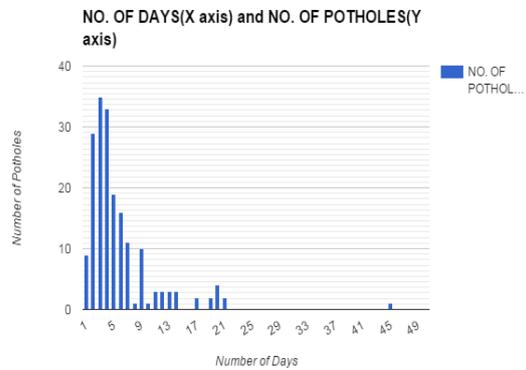

Figure 26. No large area and small depth potholes     Figure 27. Number of potholes vs days

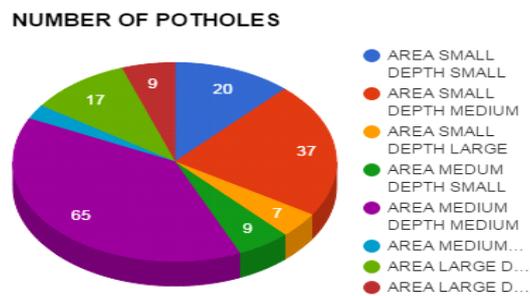

Figure 28. Pie chart depicting categories of potholes and their numbers

## 5. CONCLUSION

Phases involving Ward selection, Centralized database generation, Geotagging and Image processing have been implemented and the target pothole area has been determined from the images. The results of for a single pothole are depicted in Figure 29. Further work of obtaining dimensions from the images and comparisons with the data procured through BMC is to be implemented.

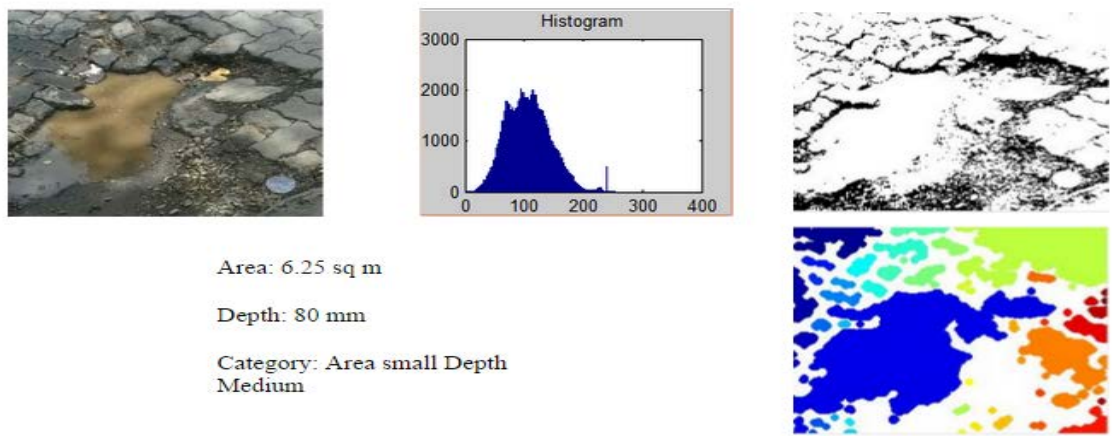

Figure 29. Results for a small area and medium depth pothole

## 6. REFERENCES

[1] Newspaper articles:
 (a)http://www.mumbaimirror.com/mumbai/cove r-story/Killer-pothole-leaves-Bandra-woman-in-coma/articleshow/46902348.cms, (b)http://www.firstpost.com/mumbai/potholes- as-death-traps-why-should-the-civic-body-go- scot-free-953561.html (c)http://indiatoday.intoday.in/story/up- engineers-contractors-to-face-culpable-homicide-charges-if-potholes-cause- death/1/486135.html (d)http://timesofindia.indiatimes.com/india/Over-11000-people-killed-by-potholes-speed- breakers-last-year/articleshow/48950267.cms

[2] Taehyeong Kim, Seung-Ki Ryu, "Review and Analysis of Pothole Detection Methods", Journal of Emerging Trends in Computing and Information Sciences, Vol. 5, No. 8 August 2014, pp. 603-608

[3] Christian Koch, Ioannis Brilakis, " Pothole detection in asphalt pavement images", Advanced Engineering Informatics 25 (2011),pp. 507–515

[4] Emir Buza, Samir Omanovic, Alvin Huseinovic, " Pothole Detection with Image Processing and Spectral Clustering",   Recent Advances in Computer Science and Networking, pp. 48-53

[5] Maher I. Rajab, Mohammad H. Alawi, Mohammed A. Saif, "Application of Image Processing  to  Measure  Road  Distresses", WSEAS Transactions on Information Science & Applications, Issue 1, Volume 5, January 2008, pp. 1-7

[6] S Nienaber, M Booysen and R Kroon, "Detecting Potholes Using Simple Image Processing Techniques And Real-World Footage", Proceedings of the 34th Southern African Transport Conference (SATC 2015), pp.153-164

[7] Sumit Pawade1, Prof. B.P. Fuladi, Prof. L.A. Hundikar, "FPGA Based Intelligent Potholes Detection System", International Journal of Innovative Research in Computer and Communication Engineering, Vol. 3, Issue 3, March 2015, pp. 2285-2290

[8] Ward:   https://en.wikipedia.org/wiki/Administrative_divisions_of_Mumbai

[9] Voice of Citizen website:   http://voiceofcitizen.com/

[10] Google Earth file:   https://drive.google.com/open?id=0B1Rifi8w3J6DMFFGeEhxX0lKQzA